\documentclass{article}
\usepackage[preprint]{neurips_2026}
\workshoptitle{Can We Trust the Judge? Building Reliable Evaluation for Language Models}
\usepackage[utf8]{inputenc}\usepackage[T1]{fontenc}\usepackage{hyperref}\usepackage{url}
\usepackage{booktabs}\usepackage{amsmath}\usepackage{amssymb}\usepackage{microtype}
\usepackage{graphicx}\usepackage{enumitem}
\title{Hard-Gate Candidacy in a Deployed Validator Suite}
\author{
  Xin Xu\\
  Carnegie Mellon University\\
  \texttt{xuxin@cmu.edu}
}
\begin{document}
\maketitle

\begin{abstract}
Before a validator can be promoted to a hard gate on a deployment pipeline, it has to be shown that its firing separates outputs that reach users in working order from those that do not. We run that screen on 13 validators in a deployed generative agent, against 550 runtime and 350 static builds labelled by downstream outcome, and report each check's marginal separation $J=\mathrm{TPR}-\mathrm{FPR}$ with Newcombe intervals and Fisher exact tests. \textbf{Two checks survive correction for multiple comparisons}, two more are nominal only, and the remaining nine are not distinguishable from zero, three of them because they never fired on any sampled build. Execution itself is not random with respect to the property being gated, and this replicates: across four runs covering 1{,}867 builds and ten distinct runtime checks, probes were skipped on \textbf{144 of 895 broken builds and 1 of 972 acceptable builds} (per-run rates 15.6\% to 16.6\% against at most 0.3\%), every skip carrying the same unsafe-to-probe reason. Because a skipped check is recorded as a pass, this imposes a ceiling that no check quality can lift: a check that needs a live artifact cannot operationally detect more than about $84\%$ of broken builds in this harness. For the one check with construct-specific labels, a detector built for blank output fires on 0 of 90 human-labelled blank builds ($95\%$ upper bound on sensitivity $3.3\%$), and the global frame statistic it approximates separates the classes only weakly (AUC $0.59$), so the gap is not a threshold that needs tuning. The same gap appears one layer up: on a census of tens of thousands of judge-scored builds, \textbf{32.5\% of rejections carry no recorded issue at all}. We argue that evaluation records must distinguish a check that ran and passed from one that did not run, must carry the evidence for a rejection, and that an inventory of checks is not evidence about a gate.
\end{abstract}

\section{Introduction}

Deployed language-model agents are wrapped in automated checking: static analyses of the generated artifact, runtime probes of its behaviour, model-based judges. These checks gate release, trigger regeneration, or route cases to human review, and the record they leave is terse---a verdict per check, summarised into a decision.

We run one specific screen on such a suite. Before a check is promoted to a hard gate, the minimum it must show is that its firing \emph{separates} the outcomes the gate exists to separate. This is deliberately not the question of whether a check detects the construct its name denotes: a check may detect its construct perfectly and still fail this screen, if that construct is uncorrelated with whether the artifact reaches users in working order. Nor is it the question of a check's incremental value inside an existing ensemble, which requires the joint firing distribution and the deployed gating rule. What we measure is the marginal screen, which is the question that was actually being asked of these checks.

The general shape of the problem is old. \citet{beer1997vacuity} formalised vacuous satisfaction in temporal-logic model checking, where a property is satisfied whenever its precondition is unsatisfiable, indistinguishable in the verdict from a meaningful pass; \citet{kupferman2003vacuity} generalise the detection semantics. Software testing developed the question independently: \citet{barr2015oracle} survey the oracle problem, \citet{schuler2011checked} measure whether executed computation reaches an oracle at all, and \citet{niedermayr2016pseudo,veraperez2019pseudo} document \emph{pseudo-tested} code, executed by a suite but removable without any test failing. What has not been measured is how this behaves in a deployed language-model-agent gate, where checks are heterogeneous, the failure base rate is low, and the artifact is generated afresh on every request.

\paragraph{Contributions.}
\begin{enumerate}[topsep=2pt,itemsep=1pt,leftmargin=*]
\item A per-check hard-gate screen of a deployed validator suite against downstream outcome, with Newcombe intervals, Fisher exact tests and Holm correction across 13 checks (\S\ref{sec:screen}). Two survive correction; nine are not distinguishable from zero, three because they never fired.
\item Evidence that execution is one-sided with respect to the gated property, recurring across four runs and ten distinct runtime checks: 15.6--16.6\% of broken builds skipped against at most 0.3\% of acceptable ones, per-run $z$ between $4.6$ and $7.8$, with skips recorded as passes---imposing a harness-level ceiling on operational sensitivity, $\mathrm{TPR}_{\mathrm{op}}\le1-s_1\approx0.84$, on every check that requires a live artifact, a mechanism we name \emph{failure-correlated execution censoring} (\S\ref{sec:applic}).
\item A construct-aligned measurement for the one check where target labels exist: 0 of 90 human-labelled instances detected, and evidence that the global frame statistic the detector approximates is itself a weak discriminator (\S\ref{sec:validity}).
\item A record prescription: applicability attested per check, so that ``ran and passed'' and ``did not run'' cease to be the same symbol (\S\ref{sec:records}).
\end{enumerate}

\section{Setting, Criterion, and Outcome Proxy}
\label{sec:setting}

\paragraph{System Under Study.} A deployed generative agent produces interactive artifacts from natural-language requests. Each request yields a build; a vision-based judge scores it and may trigger a retry. Independently of the judge, automated validators run in two lanes: \emph{static} checks inspect the generated source, \emph{runtime} checks load the build in a headless browser and probe it. The suite is larger than the subset we screen; we report the 13 checks for which both outcome classes could be labelled.

\paragraph{Screening Criterion.} We screen each check for hard-gate candidacy: does its firing separate the outcome classes? This is the question that was being asked of these checks at the time, the decision under consideration being which were clean enough to enforce. A check that detects its construct perfectly but whose firing does not separate the classes fails this screen, and that is a fact about its candidacy as a gate, not a mismeasurement of the check. We do not measure incremental contribution inside the deployed ensemble; that would require the joint firing distribution and the gating rule (\S\ref{sec:limits}).

\paragraph{Outcome Proxy.} A build is \textbf{broken} if it failed the judge \emph{or} a user reported it as not working; \textbf{acceptable} if it passed the judge, was published, and drew no complaint. These are downstream proxies, not verified ground truth, and they incorporate the judge itself; \S\ref{sec:limits} states what that costs. They are independent of any individual validator's verdict, which is what a per-check screen requires.

\paragraph{Sample Composition.} The runtime lane used 550 production builds (235 broken, 315 acceptable; broken share $0.427$ by construction). The static lane used a balanced 350-build sample (175/175). Sampling is by convenience within a single short window.

\paragraph{Statistical Inference.} We summarise separation by $J=\mathrm{TPR}-\mathrm{FPR}$, which is invariant to the prior probability of failure. $J$ weights a point of TPR equally against a point of FPR, which a cost-sensitive gate would not; it is a screen for whether a check separates the classes at all, not an operating-point recommendation. Intervals are Newcombe hybrid-score intervals on the difference of two proportions, which remain finite at zero cells where a Wald interval degenerates. Because 13 checks are screened in parallel we also report Fisher exact two-sided $p$-values with Holm adjustment. Rates are \textbf{operational}: a skipped check is counted as not firing, because that is what the pipeline records.

\section{The Hard-Gate Screen}
\label{sec:screen}

\begin{figure}[t]
\centering
\includegraphics{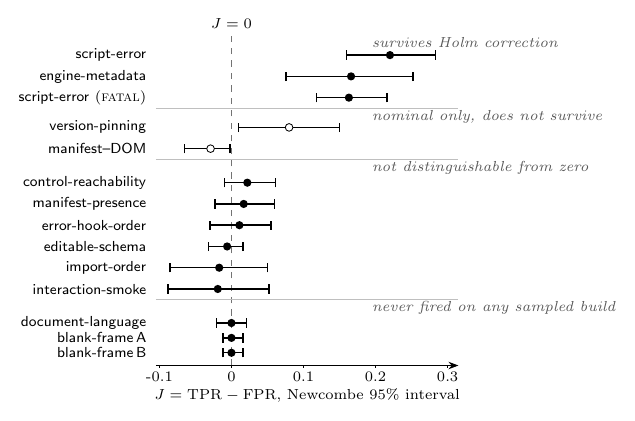}
\caption{Marginal separation $J=\mathrm{TPR}-\mathrm{FPR}$ per check with Newcombe $95\%$ intervals. Filled markers survive Holm correction or are clearly unresolved; hollow markers are nominal-only results that do not survive. The three never-fired checks have finite, not degenerate, intervals.}
\label{fig:forest}
\end{figure}

\begin{table}[t]
\centering\footnotesize
\caption{Hard-gate screen. Counts are builds on which the check fired. Rates are operational, with skips counted as non-firing. $J=\mathrm{TPR}-\mathrm{FPR}$ with Newcombe $95\%$ interval; $p$ is Fisher exact two-sided; Holm correction is over the 13 distinct checks. \textsc{fatal} is \textsf{script-error} restricted to uncaught exceptions, an ablation, and is excluded from the correction.}
\label{tab:main}
\begin{tabular}{llrrrlrr}
\toprule
Check & Lane & broken & acceptable & $J$ & Newcombe CI & $p$ & Skip \\
\midrule
\textsf{script-error} & runtime & 63/235 & 15/315 & $+0.220$ & $[+.160,+.283]$ & $<10^{-4}$ & 0 \\
\textsf{engine-metadata} & static & 57/175 & 28/175 & $+0.166$ & $[+.076,+.252]$ & $0.0004$ & 0 \\
\textsf{script-error} (\textsc{fatal}) & runtime & 39/235 & 1/315 & $+0.163$ & $[+.118,+.216]$ & $<10^{-4}$ & 0 \\
\midrule
\textsf{version-pinning} & static & 29/175 & 15/175 & $+0.080$ & $[+.010,+.150]$ & $0.035$ & 0 \\
\textsf{manifest--DOM} & static & 0/175 & 5/175 & $-0.029$ & $[-.065,-.002]$ & $0.061$ & 0 \\
\midrule
\textsf{control-reachability} & runtime & 12/235 & 9/315 & $+0.022$ & $[-.010,+.061]$ & $0.185$ & 40 \\
\textsf{manifest-presence} & static & 171/175 & 168/175 & $+0.017$ & $[-.023,+.060]$ & $0.542$ & 0 \\
\textsf{error-hook-order} & static & 170/175 & 168/175 & $+0.011$ & $[-.030,+.055]$ & $0.771$ & 0 \\
\textsf{editable-schema} & static & 174/175 & 175/175 & $-0.006$ & $[-.032,+.016]$ & $1.000$ & 0 \\
\textsf{import-order} & static & 18/175 & 21/175 & $-0.017$ & $[-.085,+.050]$ & $0.735$ & 0 \\
\textsf{interaction-smoke} & runtime & 50/235 & 73/315 & $-0.019$ & $[-.088,+.052]$ & $0.607$ & 40 \\
\midrule
\textsf{document-language} & static & 0/175 & 0/175 & $0.000$ & $[-.021,+.021]$ & $1.000$ & 0 \\
\textsf{blank-frame\,A} & runtime & 0/235 & 0/315 & $0.000$ & $[-.012,+.016]$ & $1.000$ & 40 \\
\textsf{blank-frame\,B} & runtime & 0/235 & 0/315 & $0.000$ & $[-.012,+.016]$ & $1.000$ & 40 \\
\bottomrule
\end{tabular}
\end{table}

\paragraph{Checks Surviving Multiplicity Correction.} \textsf{script-error} ($p<10^{-4}$) and \textsf{engine-metadata} ($p=0.0004$, Holm-adjusted $0.005$) separate the classes. The larger, $J=0.220$, is modest in absolute terms: the check fires on roughly one broken build in four. \textsf{version-pinning} is nominally positive ($p=0.035$) but Holm-adjusted reaches only $0.39$; \textsf{manifest--DOM} has a Newcombe interval excluding zero but Fisher $p=0.061$, so its sign is not robust to the choice of test, let alone to correction. We report both as nominal results rather than findings.

\paragraph{Checks Without Detectable Separation.} Six checks are not distinguishable from zero, by two different routes. \textsf{manifest-presence}, \textsf{error-hook-order} and \textsf{editable-schema} fire on 96--100\% of \emph{both} classes---\textsf{editable-schema} on 174 of 175 broken builds and all 175 acceptable ones---so their firing is close to constant and cannot carry information about the class. \textsf{control-reachability}, \textsf{import-order} and \textsf{interaction-smoke} fire on a minority of both classes at rates too close to separate at this sample size. A larger sample could resolve the second group; the first cannot, because a check that fires on everything has no room to discriminate.

\paragraph{Checks With No Observed Firing.} Three checks never fired: \textsf{document-language} in 350 static builds, and \textsf{blank-frame\,A} and \textsf{blank-frame\,B} in every runtime build on which they ran. Their intervals are narrow but finite: the absence of observed firing bounds the rate, it does not establish that the rate is zero. \S\ref{sec:validity} supplies the construct-aligned measurement for one of them.

\paragraph{Separation Versus Alarm Precision.} At a failure prevalence of $p=0.1$, even \textsf{script-error} would flag more false alarms than true ones ($4.3\%$ against $2.7\%$ of builds) despite the strongest $J$ in the suite---the standard low-prevalence arithmetic \citep{axelsson2000baserate}. This bears on precision, not on the sign of $J$, which is class-conditional.

\section{One-Sided Execution}
\label{sec:applic}

Every runtime check except \textsf{script-error} was skipped on 40 of 550 builds; the static lane, needing only source text, was skipped on none.

\paragraph{One-Sided Skipping Across Four Runs.} Skipping is also one-sided, and this replicates. Table~\ref{tab:skip} gives per-class skip counts for four independent runs on the same pipeline, recovered from the per-class disposition of every build. The pattern is the same in all four: probes are skipped on roughly one broken build in six and on essentially no acceptable build. Taken together the four runs cover 1{,}867 sampled builds and give \textbf{144 of 895 broken against 1 of 972 acceptable}; we report the per-run statistics as primary, because the runs draw from one harness within one window and we cannot establish that they sample disjoint builds, so a pooled test would treat correlated samples as independent. Every skip carries the same recorded \emph{unsafe-to-probe} reason. Run~C is the strongest evidence that the mechanism belongs to the harness rather than to any particular check: its four probes are disjoint from those screened in Table~\ref{tab:main}, and it shows the same rate.

\begin{table}[h]
\centering\footnotesize
\caption{One-sided execution across four runs. Counts are builds on which the probe did not execute. Run~B is the sample screened in Table~\ref{tab:main}; run~C covers four runtime checks disjoint from it. $^{\dagger}$Runs share one harness and window and may sample overlapping builds, so no pooled test statistic is reported.}
\label{tab:skip}
\begin{tabular}{llrrrr}
\toprule
Run & runtime checks & skipped / broken & skipped / acceptable & $\Delta$ & $z$ \\
\midrule
A & 4 & 29/175 \;\; 16.6\% & 0/174 \;\; 0.0\% & $+16.6$ & $5.6$ \\
B & 4 & 39/235 \;\; 16.6\% & 1/315 \;\; 0.3\% & $+16.3$ & $7.3$ \\
C & 4 (disjoint) & 20/125 \;\; 16.0\% & 0/124 \;\; 0.0\% & $+16.0$ & $4.6$ \\
D & 2 & 56/360 \;\; 15.6\% & 0/359 \;\; 0.0\% & $+15.6$ & $7.8$ \\
\midrule
combined$^{\dagger}$ & 10 distinct & \textbf{144/895 \;\; 16.1\%} & \textbf{1/972 \;\; 0.1\%} & $+16.0$ & --- \\
\bottomrule
\end{tabular}
\end{table}

This determines how Table~\ref{tab:main} reads. Our rates count a skip as a non-firing, which is what the pipeline records. Writing $s_1$ for the skip rate on broken builds and $s_0\approx0$ for acceptable ones (at most 0.3\% in any run), $\mathrm{TPR}_{\text{op}}=(1-s_1)\mathrm{TPR}_{\text{run}}$ and $\mathrm{FPR}_{\text{op}}\approx\mathrm{FPR}_{\text{run}}$, so $J_{\text{run}}-J_{\text{op}}\approx s_1\mathrm{TPR}_{\text{run}}\ge0$: conditional-on-running values are \emph{weakly} higher, strictly so only for a check that fires on runnable broken builds at all; for \textsf{blank-frame\,A} and \textsf{blank-frame\,B} the two coincide at zero. \textbf{A verdict-only record cannot distinguish them.}

\paragraph{A Harness-Level Ceiling on Operational Sensitivity.} Since $\mathrm{TPR}_{\text{run}}\le1$, the same identity yields $\mathrm{TPR}_{\text{op}}\le1-s_1$: in this harness, \textbf{no check that requires a live artifact can operationally detect more than $0.83$--$0.84$ of broken builds}, whatever its logic, because the harness withholds exactly the builds such checks most need to see. We call the mechanism \emph{failure-correlated execution censoring}: the probability that a check executes falls with the severity of the condition the check exists to detect---the opposite pole to gray failure \citep{huang2017grayfailure}, where the degradation is too subtle for the detector rather than too severe for it to run. The ceiling is a property of harness policy, not of any check, and a verdict-only record is precisely what lets it persist unnoticed. Consistent with it, the one runtime check that executed on every build, \textsf{script-error}---it observes the load itself rather than probing the loaded page---is also the only runtime check whose separation survives correction.

\section{Construct Validity Where Target Labels Exist}
\label{sec:validity}

For one check we can measure validity as well as separation. \textsf{blank-frame\,A} and \textsf{blank-frame\,B} exist to detect blank or single-colour output, a failure users describe directly. Within the 550-build screen, 14 builds were independently confirmed blank; the detectors fired on none.

To go beyond fourteen cases, we obtained construct-specific labels: 239 builds captured as screenshots and labelled blank or acceptable by human review, 90 blank and 149 not. The detector fired on \textbf{0 of 90} blank builds and 0 of 149 acceptable ones, an exact binomial $95\%$ upper bound on sensitivity of $\mathbf{3.3\%}$. We report this set separately from the 14 audit cases rather than pooling them, because we cannot establish that the two sets are disjoint.

\paragraph{Threshold Tuning Versus Construct Redesign.} The gap is not a threshold that needs tuning. The shipped detector samples a sparse grid of points and reports a blank frame only when every sampled point has nearly the same colour. Blank builds do have lower median frame luminance than acceptable ones, 37.1 against 57.3, but the distributions overlap heavily: as a single feature, mean frame luminance has $\mathrm{AUC}=0.59$ ($95\%$ CI $[0.52,0.67]$), and its best single threshold reaches only $\mathrm{TPR}=0.84$ at $\mathrm{FPR}=0.66$. Human raters separate these classes; a global frame statistic does not. Repairing this detector therefore means changing what it measures, not where it cuts---a different and more expensive fix than a verdict-only record would reveal was needed.

\section{Implications for Evaluation Records}
\label{sec:records}

The suite records per-build verdicts. That record cannot express any of the states above. A build for which every check returned \emph{pass} is written identically whether the checks ran and found nothing, did not run, never fire, or fire on everything. The aggregate claim---the suite passed it---is not falsifiable from the record.

\paragraph{Record Completeness at the Judge Layer.} The same gap appears one layer up, at scale. The vision judge records, per build, a verdict and a list of issues. On a census of every scored build in the same window, tens of thousands in all, these two are not consistent with each other in a specific direction: \textbf{32.5\% of rejections (Wilson $95\%$ $[31.1,34.0]$) carry no recorded issue at all}, while acceptance with a recorded \textsc{critical} issue is rare ($0.1\%$ of accepted builds). The record therefore does not reconstruct the verdict on the reject side: in roughly a third of rejections, nothing in the artifact explains why the build was rejected. We claim this about the record, not the judge's reasoning: a downstream consumer---a human triaging failures, an auditor, a retraining pipeline---has nothing to work from in those cases.

The remedy is narrow. A record should carry, per check and per item, whether the check was \emph{applicable}, so that the record is three-valued---fired, ran without firing, did not run---rather than binary; diagnostic-accuracy reporting has long treated inconclusive results this way rather than folding them into a negative \citep{shinkins2013inconclusive}. At the suite level, a periodic screen of the kind in Table~\ref{tab:main}, with intervals and multiplicity control, converts an inventory of checks into a statement about what the suite can separate. Both require only information already present in the pipeline, which is currently discarded when it is summarised.

This matters because validator inventories are cited as evidence in deployment and assurance arguments. \citet{bucknall2024openproblems} identify insufficient external access as the limiting factor for governance-relevant evaluation, and single out downstream user logs as a further-restricted gap. Our results add a complication that access alone does not resolve: with full internal access, nine of thirteen checks showed no separation distinguishable from zero, two more were nominal-only, and four were in addition skipped on the builds that most needed checking.

\section{Related Work}

\paragraph{Vacuity, Test Oracles, and Pseudo-Tested Code.} \citet{beer1997vacuity} and \citet{kupferman2003vacuity} established that a passing verdict may carry no information. Software testing developed the question independently: \citet{barr2015oracle} survey the oracle problem, \citet{schuler2011checked} propose checked coverage, and \citet{niedermayr2016pseudo,veraperez2019pseudo} document pseudo-tested code at scale. Our never-firing and constant-firing states are the deployed-gate analogue of that phenomenon. What differs is the instrument: that line establishes the property \emph{interventionally}, by mutating source or injecting faults and observing whether any check reacts \citep{hsueh1997faultinjection,jahangirova2016oracle}, which presupposes the access and the harness control to do so; we establish it \emph{observationally}, from production outcomes on a running gate. The two designs answer different questions: intervention asks whether a check can react to a seeded fault, observation asks whether its firing separates the outcomes that production actually delivers.

\paragraph{Reliability of Judges and Guardrails.} \citet{catchingonefive2026} audit a judge gate on a deployed agent and find pooled recall near 18\%; \citet{bellso2026} benchmark 28 supervision systems on in-house data; \citet{hallmark2026} argue that false-positive rate rather than recall decides whether a verifier is deployable; \citet{kamoi2024realmistake} establish low recall for LLM error detectors; \citet{jung2025trust} give escalation with provable agreement guarantees. An independent audit of commercial moderation APIs uses curated datasets \citep{hartmann2025moderation}, and a platform-side study on real moderation traffic withholds its class ratios \citep{son2023wild}.

\paragraph{Selective Execution and Verification Bias.} A check that runs on part of the population is a selective predictor \citep{geifman2017selective}. Diagnostic-test methodology treats accuracy under selective verification \citep{begg1983verification}, but that structure is the mirror of ours: there the index test is always observed and the reference standard is selectively missing, and the standard correction assumes selection depends on the observed test result rather than true status. Here the reference standard is complete and the \emph{index test} is missing, under a rule keyed on severity. We therefore report operational rates and bound the conditional ones rather than attempt a verification-bias correction.

\paragraph{Failures of Evaluation Pipelines.} \citet{swecycle2026} attribute close to a third of scoring disagreements to the evaluation pipeline itself; \citet{silentfailures2026} taxonomises silent failures where a monitor runs and is deceived, whereas our states concern checks that do not fire or do not run. \citet{huang2017grayfailure} define gray failure as differential observability for subtle degradations; \S\ref{sec:applic} develops the opposite pole.

\section{Limitations}
\label{sec:limits}

\textbf{Estimand.} We measure marginal separation, which is the screen for hard-gate candidacy, not incremental contribution inside the deployed ensemble. Two high-$J$ checks may be redundant; two low-$J$ checks may cover disjoint rare failures. Deciding that needs the joint firing matrix and the gating rule, which we do not have.

\textbf{Construct Scope.} Only \S\ref{sec:validity} has construct-specific labels, and only for one check. A check failing this screen may be a correct detector of a condition that does not separate our outcome classes. Establishing validity for the remaining twelve requires seeded failures or target labels; the interventional designs cited above are the right instrument for that.

\textbf{Labels and Inference.} Outcome labels derive from the deployment's own judge and from absence of complaints, with no uniform complaint window, so builds sampled near the end of the window had less exposure. Human blank labels lack a stated annotation protocol, rater count, or agreement statistic. Retries can produce several builds per request, so builds are not fully independent and our intervals do not use cluster-robust inference. The judge's issue list may likewise be incomplete, so \S\ref{sec:records} reports record completeness, not judge correctness.

\textbf{Sampling.} One deployment, convenience sampling, a single short window; the four skip runs share one harness and window and may overlap. Rates are not claimed to characterise validator suites generally.

\bibliographystyle{plainnat}

\end{document}